\documentclass[sigconf]{acmart}

\usepackage{booktabs} 
\usepackage{multirow}
\usepackage{multicol}
\usepackage{balance}
\usepackage{graphicx}
\usepackage{epstopdf}

\copyrightyear{2018}
\acmYear{2018}
\setcopyright{acmcopyright}
\acmConference[MM '18]{2018 ACM Multimedia Conference}{October 22--26,
	2018}{Seoul, Republic of Korea}
\acmBooktitle{2018 ACM Multimedia Conference (MM '18), October 22--26,
	2018, Seoul, Republic of Korea}
\acmPrice{15.00}
\acmDOI{10.1145/3240508.3240559}
\acmISBN{978-1-4503-5665-7/18/10}

\begin{document}
\title{Text-to-image Synthesis via Symmetrical Distillation Networks}

\author{Mingkuan Yuan and Yuxin Peng*}
\affiliation{%
	\institution{Institute of Computer Science and Technology, Peking University}
	\city{Beijing}
	\country{China}}
\email{pengyuxin@pku.edu.cn}

\begin{abstract}
Text-to-image synthesis aims to automatically generate images according to text descriptions given by users, which is a highly challenging task. The main issues of text-to-image synthesis lie in two gaps: the heterogeneous and homogeneous gaps. The \textbf{heterogeneous gap} is between the high-level concepts of text descriptions and the pixel-level contents of images, while the \textbf{homogeneous gap} exists between synthetic image distributions and real image distributions. For addressing these problems, we exploit the excellent capability of generic discriminative models (e.g. VGG19), which can guide the training process of a new generative model on multiple levels to bridge the two gaps. The high-level representations can teach the generative model to extract necessary visual information from text descriptions, which can bridge the \textbf{heterogeneous gap}. The mid-level and low-level representations can lead it to learn structures and details of images respectively, which relieves the \textbf{homogeneous gap}. Therefore, we propose Symmetrical Distillation Networks (SDN) composed of a source discriminative model as ``teacher'' and a target generative model as ``student''. The target generative model has a symmetrical structure with the source discriminative model, in order to transfer hierarchical knowledge accessibly. Moreover, we decompose the training process into two stages with different distillation paradigms for promoting the performance of the target generative model. Experiments on two widely-used datasets are conducted to verify the effectiveness of our proposed SDN.
\end{abstract}

\maketitle


\section{Introduction}

As a multimedia technology, cross-modal retrieval \cite{LiMM03CFA, RasiwasiaMM10SemanticCCA, zhang2014start, feng12014cross, wang2017adversarial} has become a highlighted topic of research in the big data era. However, cross-modal retrieval is unable to meet user's needs sometimes because it only can provide existing data for users. So the research on cross-modal synthesis has been drawing more attention in the community, such as text-to-image synthesis. Text-to-image synthesis can generate images from text descriptions automatically. It is a promising but challenging task, which has two gaps as the main issues. On one hand, there is a \textbf{heterogeneous gap} between the high-level concepts in text descriptions and the pixel-level contents of synthetic images. The heterogeneous gap corresponds to the semantic relevance between input text descriptions and generated images. On the other hand, there is a \textbf{homogeneous gap} between synthetic and real image distributions due to the huge image pixel space. The homogeneous gap is reflected in the authenticity and quality of synthetic images, including the global structures and local details. Therefore, the goal of text-to-image synthesis is to bridge these gaps and generate images with high authenticity, quality and semantic relevance conditioned on text descriptions.


For addressing the above issues, we consider the general paradigm of the discriminative tasks. There is a general paradigm of the discriminative tasks, which is training a generic feature representation model by a large-scale dataset of labeled images firstly, and then fine-tuning the model for the specific task \cite{yosinski2014transferable}. We can find many successful applications of this paradigm in the various state-of-the-art methods, such as object detection \cite{girshick2016region}, semantic segmentation \cite{long2015fully} and fine-grained image classification \cite{he2017fine}.

\begin{figure}[t]
	\begin{center}
		\includegraphics[width=1\linewidth]{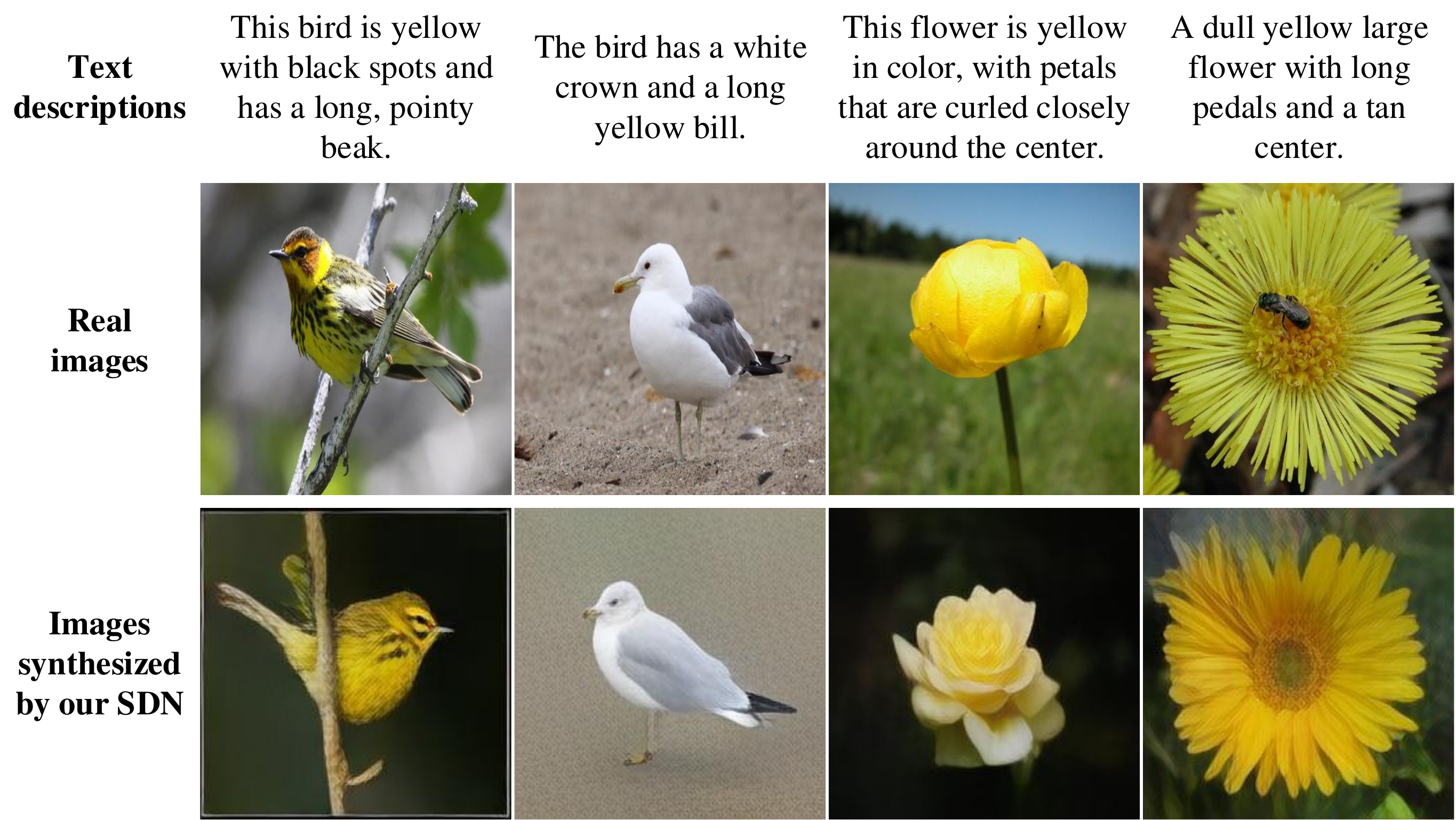}
	\end{center}
	\caption{The examples of images synthesized by our proposed SDN, compared to the real images.}
	\label{fig:example}
\end{figure}

By contrast, there is no generic model for generative tasks and most of existing generative methods based on deep neural networks train their models from scratch. However, the \textbf{generic discriminative model has a strong capability to produce multi-level representations}, which can be verified by the research on inverting and understanding it \cite{mahendran2015understanding, dosovitskiy2016inverting, zhang2016augmenting}. By inverting and understanding, it can be noted that these multi-level representations contain hierarchical information from pixel content to semantic concept. So \textbf{it is helpful to bridge the heterogeneous and homogeneous gaps of text-to-image synthesis} by using the capability of the generic discriminative model. The generic discriminative model is generally based on the image classification task and can predict the semantic labels of input images. It can map pixel-level contents of images to high-level concepts, which is a reverse process against text-to-image synthesis. So the \textbf{high-level representations} produced by the generic discriminative model can be a guidance for generative models, which teaches them to extract necessary visual information from text descriptions and bridges the heterogeneous gap. Moreover, the \textbf{mid-level and low-level representations} can lead generative models to learn structures and details of images respectively, which relieves the homogeneous gap problem. It is easier to find the optimal representations for synthetic images in these low-dimensional feature spaces than the high-dimensional image pixel space. Therefore, there are significant advantages to train the generative model of text-to-image synthesis by the generic discriminative model.

In this paper, we propose Symmetrical Distillation Networks (SDN) to fully utilize the strong capability of the generic discriminative model for the generative model of text-to-image synthesis. The ``distillation'' paradigm is inspired by the \cite{hinton2015distilling}. Hinton et al. \cite{hinton2015distilling} propose the ``distillation'' concept from the model compression idea of Bucilua et al. \cite{bucilua2006model}, which can compress large models trained on large datasets to a much smaller model. Our work further extends this idea between a generic discriminative model and a new generative model.

We design a symmetrical structure for SDN composed of a source discriminative model and a target generative model (hereinafter referred to as the ``source model'' and ``target model'' respectively). The source model is a generic discriminative model (e.g. VGG19 \cite{simonyan2014very}), and the target generative model has the same layers but reverse data flow direction with the source model. The symmetrical structure makes it accessible to distill knowledge from the source model to the target model on multiple corresponding levels.

Moreover, we design two distillation ways in different stages: Stage-I and Stage-II distillation. By \textit{Stage-I distillation}, the target model learns from representations given by source model roughly, which can draw blurry images with almost all visual information conditioned on text descriptions. By \textit{Stage-II distillation}, the target model learns more about the tiny discrepancy between synthetic and real images in multiple levels, synthesizing images with more details about the object finally.

We conduct comparative experiments with the state-of-the-art methods on two widely-used datasets to verify the effectiveness of our proposed SDN. Moreover, several baseline experiments are conducted to analyze the importance of different components.

\section{Related Work}

Due to the deep networks \cite{lecun2015deep}, image generation has dramatic progress in recent years. Although the early work such as \cite{mnih2010generating} can only produce synthetic images which are easy to distinguish from real samples, the recent work \cite{nguyen2016synthesizing, zhang2016stackgan, chen2017photographic} can synthesize photo-realistic images. There are some typical deep image generation approaches promoting this progress. Kingma et al. propose Variational Autoencoders (VAE) \cite{kingma2013auto} by using probabilistic graphical models and maximizing the lower bound of data likelihood to formulate the generation problem. Goodfellow et al. propose Generative Adversarial Networks (GAN) \cite{goodfellow2014generative} to train a generative model with a discriminative model in an adversarial paradigm. Deep Recurrent Attention Writer (DRAW) method \cite{gregor2015draw} can generate photo-realistic images with the recurrent variational auto-encoder and the attention mechanism. As an autoregressive method, PixelRNN \cite{oord2016pixel} realizes image synthesis by modeling the conditional distribution in the pixel space. In addition, it has been proven that conditional image generation based on these generative approaches can be realized, which has more flexible application \cite{van2016conditional, yan2016attribute2image}.

In the last few years, more and more researchers focus on the text-to-image synthesis problem, which is an image generation task conditioned on text descriptions. Since the main issues of text-to-image synthesis lie in the homogeneous and heterogeneous gaps, existing methods have considered these problems and attempt to deal with them. The earlier work focuses more on the \textbf{heterogeneous gap}.
Mansimov et al. \cite{mansimov2015generating} introduce an AlignDRAW model to estimate alignment between generated images and text descriptions by a soft attention mechanism. However, the attention mechanism ignores some visual detailed information of text descriptions.
Reed et al. \cite{reed2016generative} combine deep symmetric structured joint embedding \cite{reed2016learning} and GAN \cite{goodfellow2014generative}. The embedding approach can excavate more detailed visual information for generated images, but not fully utilize these information due to the elementary heterogeneous correlation constraints. 
To bridge the \textbf{homogeneous gap}, Reed et al. \cite{reed2016gawwn} propose GAWWN to control the global structures by object and part keypoint location constraints simultaneously. This method can generate images with correct object shapes and colors successfully, but still lack the authenticity in details. To relieve the homogeneous gap problem, StackGAN \cite{zhang2016stackgan} splits the text-to-image synthesis task into two stages. StackGAN generates images via the rough structures in the first stage and it uses these images to further generate images with more local details in the second stage. Limited by first stage, the structures of final generated images have some unrealistic characteristics and the homogeneous gap still exists.

Instead of adversarial training widely used in previous methods, the generative model of our proposed SDN has a feedforward structure and learns under the guidance of generic discriminative model. With this paradigm, SDN does not need minimax optimization and careful parameter adjustment of two adversarial models, which is more stable than contemporary GAN based methods \cite{arjovsky2017towards}.

\section{Symmetrical Distillation Networks}

\begin{figure*}
	\begin{center}
		\includegraphics[width=1\linewidth]{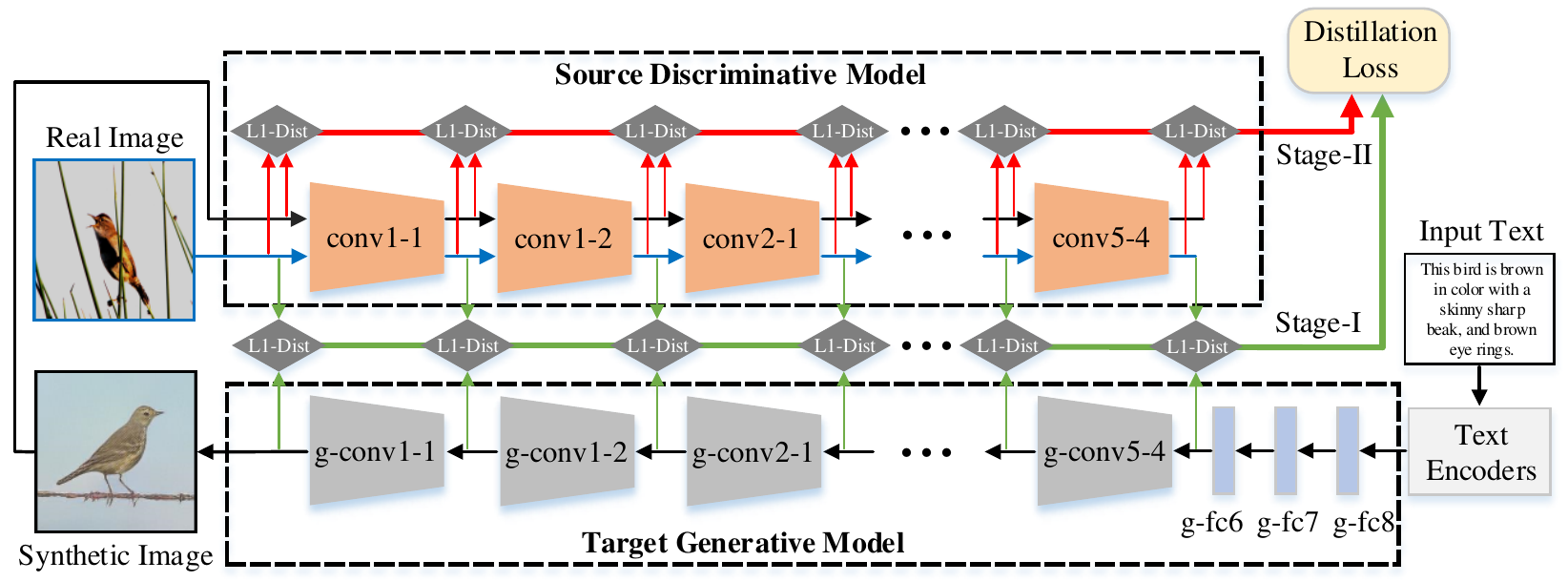}
	\end{center}
	\caption{The architecture of proposed Symmetrical Distillation Networks (SDN), which consists of a source discriminative model and a target generative model. The source model receives images as input and produces multi-level representations as guidance for the training of target model. The target model generates images conditioned on the text embedding produced by text encoders. The SDN applies two kinds of distillation loss in different stage to transfer hierarchical knowledge from the source model to the target model.}
	\label{fig:framework}
\end{figure*}

As shown in Figure~\ref{fig:framework}, our SDN is composed of a source discriminative model and a target generative model, which has a symmetrical structure. In training phase, there are two distillation stages to transfer knowledge from source model to target model by multi-level representations. In testing phase, SDN can synthesize images with multi-level characteristics of the object finally. Moreover, since a given text description can correspond to many images, we extend our SDN to a diverse paradigm. 

\subsection{Preliminaries}
Given a text description $t$, we first encode it as a text embedding $\varphi_t$ by deep convolutional and recurrent text encoders proposed by \cite{reed2016learning}. Consider the target model $G$, the parameters of $G$ is $\theta$ and $\theta_l$ denotes the parameters of a certain layer $l$. The representation produced by $l$ in $G$ can be denoted as $\phi_l^G(\varphi_t;\theta_l)$. The synthetic image is $I_s(\varphi_t;\theta) \in \mathbb{R}^{m \times n \times 3}$, where $m \times n$ refers to the image pixel resolution, and $3$ indicates the three channels of an image. Then we use its paired real image $I_r \in \mathbb{R}^{m \times n \times 3}$ as a groundtruth. Our goal is to train $G$ to generate synthetic images $I_s(\varphi_t;\theta)$ conditioned on $t$ by using the real image $I_r$ and make $I_s(\varphi_t;\theta)$ photo-realistic and related with $t$. Moreover, given an image $I$ as input, we denote the representation produced by one certain layer $l$ in the source model $D$ as $\phi_l^D(I)$.

\subsection{Architecture}

The source model $D$ in SDN is a generic feature representation model and we employ VGG19 \cite{simonyan2014very} in our experiments due to its capability and versatility. The original VGG19 model has 16 convolutional layers (conv1-1, conv1-2, conv2-1, ..., and conv5-4) with 3$\times$3 filters and 3 fully-connected layers (fc6, fc7, and fc8). Since the 3 fully-connected layers are usually regarded as task-specific layers while 16 convolutional layers are generic for all kinds of discriminative tasks, we only use these convolutional layers for distillation to avoid unnecessary interference. So $D$ is composed of 16 convolutional layers from VGG19, which can be divided into 5 groups according to the down-sampling processing. This model receives image $I$ as input and its each layer $l$ can produce multi-level representations $\phi_l^D(I)$. Moreover, considering that our experiments use a zero-shot setting, the semantic labels are not used for fine-tuning $D$ to maintain its versatility and avoid over-fitting situation.

The target model $G$ in SDN has a similar structure with the source model, consisting of 3 fully-connected layers (g-fc8, g-fc7, and g-fc6) and 16 convolutional layers (g-conv5-4, g-conv5-3, g-conv5-2, ..., and g-conv1-1) with 3$\times$3 filters. These convolutional layers can also be divided into 5 groups according to the up-sampling processing. The up-sampling processing can resize the input from a low-dimensional matrix to a high-dimensional matrix. The name of $G$ layers is homologous as VGG19 layers. This model receives the embedding $\varphi_t$ of text description $t$ as input and generates image $I_s$. Each layer $l$ of $G$ can also produce multi-level representations $\phi_l^G(\varphi_t;\theta_l)$, whose dimension is the same as the corresponding $D$ representation $\phi_l^D(I)$. So the resolution of generated images is 224$\times$224, which is the same as the dimension of $D$ input layer. 

\subsection{Two-stage Distillation}

Since the conditional distribution of the pixel space is hard to fit, it is critical to establish an appropriate loss function. It is not reasonable to only use the pixel colors of real images as goundtruth to train the generative model, which does not truly give the generative model comprehension of the correlation between images and text descriptions. For example, if the model can generate a yellow flower successfully from the description, it will fail to generate a blue flower with the same shape, because the difference between these images are huge in pixel space. So we design a distillation paradigm in multiple levels inspired by \cite{gupta2016cross} and two kinds of perceptual loss function by \cite{chen2017photographic}, which are optimized in two stages individually for different purposes.

\subsubsection{Stage-I Distillation}
The green line labeled ``Stage-I'' in Figure~\ref{fig:framework} refers to the main process of Stage-I distillation. When an embedding $\varphi_t$ of text description $t$ is sent into the target model $G$ and its paired real image $I_r$ is input into the source model $D$, $G$ and $D$ will produce corresponding representations $\phi_l^G(\varphi_t;\theta_l)$ and $\phi_l^D(I_r)$, where $\theta_l$ denotes parameters of $G$ the layer $l$. Our proposed scheme uses discrepancy between these corresponding representations as the multi-level difference between real and synthetic images. So we regard this discrepancy as the loss function in first distillation stage as follows:
\begin{align}
	\begin{split}
		Loss_{I} = \sum_l || \phi_l^D(I_r) - \phi_{l+1}^G(\varphi_t;\theta_l) ||_1 + Loss_{Image}
	\end{split}
\end{align}
In this equation, $||\cdot||_1$ denotes the L1 distance.  $ Loss_{Image}$ is the image space loss, which is a foundational constraint to learn the distribution of real images for $G$ and can be presented as follows:
\begin{align}
	Loss_{Image} = || I_r - I_s(\varphi_t;\theta) ||_1
\end{align}

The purpose of Stage-I distillation is to train $G$ by learning rich representations for real image in multi-level feature space directly. By optimizing this loss function, each layer of $G$ will be able to produce a similar feature representation $\phi_l^G(\varphi_t;\theta_l)$ with $\phi_l^D(I_r)$.

\subsubsection{Stage-II Distillation}
The red line labeled ``Stage-II'' in Figure~\ref{fig:framework} refers to Stage-II distillation progress. After the training of Stage-I distillation, the target model $G$ has the capability to generate synthetic images $I_s(\varphi_t;\theta)$ roughly from text descriptions embedding $\varphi_t$. So we send the synthetic image $I_s(\varphi_t;\theta)$ into the source model $D$ to get its multi-level representations $\phi_l^D(I_s(\varphi_t;\theta))$, which represent the cognition of $D$ to $I_s(\varphi_t;\theta)$. The cognition discrepancy of $D$ between real image $I_r$ and synthetic image $I_s(\varphi_t;\theta)$ is more detailed in multiple levels. We use this discrepancy as the Stage-II distillation as follows:
\begin{align}
	\begin{split}
		Loss_{II} = \sum_l || \phi_l^D(I_r) - \phi_l^D(I_s(\varphi_t;\theta)) ||_1 + Loss_{Image}
	\end{split}
\end{align}

If we optimize this loss function and find the optimal parameters $\theta$, $G$ will learn more about the tiny discrepancy between synthetic and real images in multiple levels, capturing integral text information.

\subsection{Diverse Synthesis}
Given a text description embedding $\varphi_t$ as input, we can get a single synthesized image by the generative model of SDN trained by architecture and scheme described above. However, the text-to-image synthesis is a one-to-many problem: a given text description can correspond to many generated images. So it is reasonable to generate diverse images for one text description as output. We change channels of $G$ final layer from $3$ to $3n$ as \cite{chen2017photographic}, where $n$ is the number of diverse images and each consecutive 3 channels forms an image. Besides, we need to change the loss function for diverse synthesis because directly using above loss functions on this modified architecture will make the diverse outputs exactly the same. Therefore, we adopt multiple choice learning \cite{guzman2012multiple} to modify the distillation loss in two stages as follows:

\begin{align}
	\begin{split}
		Loss_{I}^{Div} = & \sum_l || \phi_l^D(I_r) - \phi_{l+1}^G(\varphi_t;\theta_l) ||_1 \\
		& + \min_p || I_r - I^p_s(\varphi_t;\theta) ||_1
	\end{split}
\end{align}
\begin{align}
	\begin{split}
		Loss_{II}^{Div} = \min_p [ & \sum_l || \phi_l^D(I_r) - \phi_l^D(I^p_s(\varphi_t;\theta)) ||_1 \\
		& + || I_r - I^p_s(\varphi_t;\theta) ||_1 ]
	\end{split}
\end{align}
where $I^p_s(\varphi_t;\theta)$ denotes $p$-th synthetic image output by the $G$ final layer.

This loss function concentrates on the best synthesized image and gains its gradient to update all the output channels. By optimizing this diversity loss function, we can get a generative model that synthesizes diverse images in application.

\subsection{Implementation Details}
As shown in Figure~\ref{fig:framework}, in the source model, there are two pathways: real image pathway and synthetic image pathway. We take 16 convolutional layers and 5 down-sampling layers of VGG19 \cite{simonyan2014very} pre-trained on ImageNet dataset \cite{krizhevsky2012imagenet}. It receives the resized 224$\times$224 images as input and generates multi-level convolutional feature maps for images. Due to the fact that this model is regarded as a generic model, its parameters are frozen in training phase.

The target model has a similar structure as the source model consisting of 16 convolutional layers, 5 up-sampling layers and 3 fully-connected layers. The dimension of each layer is decided by the corresponding layer in the source model for calculating distillation loss function. The base learning rate is set to be $10^{-4}$ and we use the Adam optimization method \cite{kingma2014adam} in TensorFlow \footnote{https://www.tensorflow.org} to optimize the loss function. We train 100 epochs of Stage-I distillation and then 100 epochs of Stage-II distillation to train the generative model by a minibatch size of 12. In testing phase, it can synthesize images with 224$\times$224 resolution, which is same as the resized input real images. Moreover, we implement the up-sampling layers in the generative model via ``image resize'' operation in Tensorflow, which refers to StackGAN [32]. The convolutional layer can be combined with the up-sampling layer to realize the same effect of the deconvolutional layer.

\section{Experiments}


\subsection{Datasets}
We adopt two widely-used datasets to conduct our experiments: CUB-200-2011 \cite{wah2011caltech} and Oxford-Flower-102 \cite{nilsback2008automated} datasets. CUB-200-2011 dataset contains 11,788 images from 200 bird categories. Different with \cite{zhang2016stackgan}, we use the original images instead of cropping them in training phase because the background is also an important visual element for photo-realistic image synthesis. Oxford-Flower-102 dataset contains 102 flower categories with 8,189 images and we also use the original images same as the \cite{zhang2016stackgan}. There are 10 captions for every image in CUB-200-2011 and Oxford-Flower-102 datasets provided by \cite{reed2016learning}. Following \cite{reed2016generative} and \cite{zhang2016stackgan}, we use a zero-shot experimental setting, splitting each dataset into class-disjoint training set and testing set. CUB-200-2011 dataset is split into 150 training categories and 50 testing categories, while Oxford-Flower-102 is split into 82 training categories and 20 testing categories.

\subsection{Quantitative Evaluation}  

For quantitative evaluation, we adopt four kinds of metrics: \textbf{inception score} \cite{salimans2016improved}, \textbf{SSIM}\cite{wang2004image}, \textbf{FSIM}\cite{zhang2011fsim} and \textbf{human rank}.

Inception score is a numerical assessment approach, which has been widely used in image generation \cite{salimans2016improved, huang2017stacked, gulrajani2017improved, berthelot2017began, ma2017pose}. It can be denoted as:
\begin{align}
	I = \exp(\mathbb{E}_xD_{KL}(p(y|x)||p(y)))
\end{align}
where $x$ is a generated image, and $y$ is the label predicted by the Inception model \cite{szegedy2016rethinking}. Better methods should generate more meaningful and diverse images. So the larger KL divergence $D_{KL}(p(y|x)||p(y))$ is better. Following \cite{salimans2016improved, huang2017stacked}, we use the pre-trained Inception model \footnote{http://download.tensorflow.org/models/image/imagenet/inception-2015-12-05.tgz} and generate a large number of images to evaluate each method.

For the inception score evaluation, we carefully use the code and trained models provided by their authors to generate a large number of images (29330 on CUB-200-2011 dataset and 11550 on Oxford-Flower-102 dataset) for fair comparison. \textbf{Since the training of GAWWN needs annotations of object bounding box and part keypoints that Oxford-Flower-102 dataset does not provide, we only compare the results of our SDN with GAN-INT-CLS and StackGAN on the Oxford-Flower-102 dataset.} 

SSIM and FSIM are two image quality assessment methods, which can measure similarity between two paired images. We select paired text descriptions of all the images in testing set and generate images by compared methods for them. Then we compute the SSIM and FSIM scores between each real and generated images pairs and report the average values in the following table. The inception, SSIM and FSIM scores of our proposed SDN and compared methods are reported in Table~\ref{table:Inception}. \textbf{In this table, the higher scores are better}.

\begin{table}[htb]
	\caption{Inception, SSIM and FSIM scores of our SDN and compared methods. Higher scores mean better results. }
	\begin{center}
		\begin{tabular}{|c|c|c|c|c|c|c|} 
			\hline
			Datasets & Methods & Inception & SSIM & FSIM\\
			\hline
			
			\multirow{4}{1.5cm}{CUB-200-2011} & \textbf{our SDN} & \textbf{6.89 $\pm$ 0.06} & \textbf{0.3160} & \textbf{0.6264} \\
			& StackGAN & 4.95 $\pm$ 0.04 & 0.2812 & 0.5869 \\
			& GAWWN & 5.22 $\pm$ 0.08 & 0.2370 & 0.5653 \\
			& GAN-INT-CLS & 5.08 $\pm$ 0.08 & 0.2934 & 0.6082 \\
			\hline
			
			\multirow{3}{1.5cm}{Oxford-Flower-102} & \textbf{our SDN} & \textbf{4.28 $\pm$ 0.09} & \textbf{0.2174} & \textbf{0.6227} \\
			& StackGAN & 3.54 $\pm$ 0.07 & 0.1837 & 0.6009 \\
			& GAN-INT-CLS & 4.17 $\pm$ 0.07 & 0.1948 & 0.6214 \\
			\hline
		\end{tabular} 
	\end{center}
	\label{table:Inception}
\end{table}

Our SDN achieves the best results on two datasets compared with the other methods, verifying that it is good at bridging homogeneous gaps.

Moreover, we conduct two human rank evaluations. The \textbf{quality rank} means that users are asked to rank the generated images according to the image quality, which reflects whether the homogeneous gap is well relieved. The \textbf{correlation rank} is according to the correlation between generated images and text descriptions, corresponding to the heterogeneous gap. We select 5000 text descriptions randomly for each dataset and generate images by compared methods for each sentence, which 10 users (not including the authors) are asked to rank. The rank results are reported in Table~\ref{table:Human}. \textbf{Smaller rank values mean better results}.

\begin{table}[htb]
	\caption{Human ranks of our SDN and compared methods. Smaller rank values mean better results.}
	\begin{center}
		\begin{tabular}{|c|c|c|c|}
			\hline
			\multirow{2}{*}{Datasets} & \multirow{2}{*}{Methods} & \multicolumn{2}{|c|}{Human ranks} \\
			\cline{3-4}
			& & Quality & Correlation \\
			\hline
			\multirow{4}{1.5cm}{CUB-200-2011} & \textbf{our SDN} & \textbf{2.26 $\pm$ 1.03} & \textbf{2.23 $\pm$ 1.01} \\
			& StackGAN & 2.44 $\pm$ 1.12 & 2.44 $\pm$ 1.13 \\
			& GAWWN & 2.53 $\pm$ 1.10 & 2.48 $\pm$ 1.10 \\
			& GAN-INT-CLS & 2.77 $\pm$ 1.16 & 2.83 $\pm$ 1.15 \\
			\hline
			
			\multirow{3}{1.5cm}{Oxford-Flower-102} & \textbf{our SDN} & \textbf{1.74 $\pm$ 0.77} & \textbf{1.77 $\pm$ 0.78}  \\
			& StackGAN & 1.87 $\pm$ 0.75 & 1.90 $\pm$ 0.76 \\
			& GAN-INT-CLS & 2.39 $\pm$ 0.79 & 2.33 $\pm$ 0.81 \\
			\hline
		\end{tabular} 
	\end{center}
	\label{table:Human}
\end{table}

Our SDN also achieves the best average human rank on both datasets and both evaluation schemes, indicating that it is effective to bridge homogeneous and heterogeneous gaps. 

\begin{figure*}
	\begin{center}
		\includegraphics[width=0.85\linewidth]{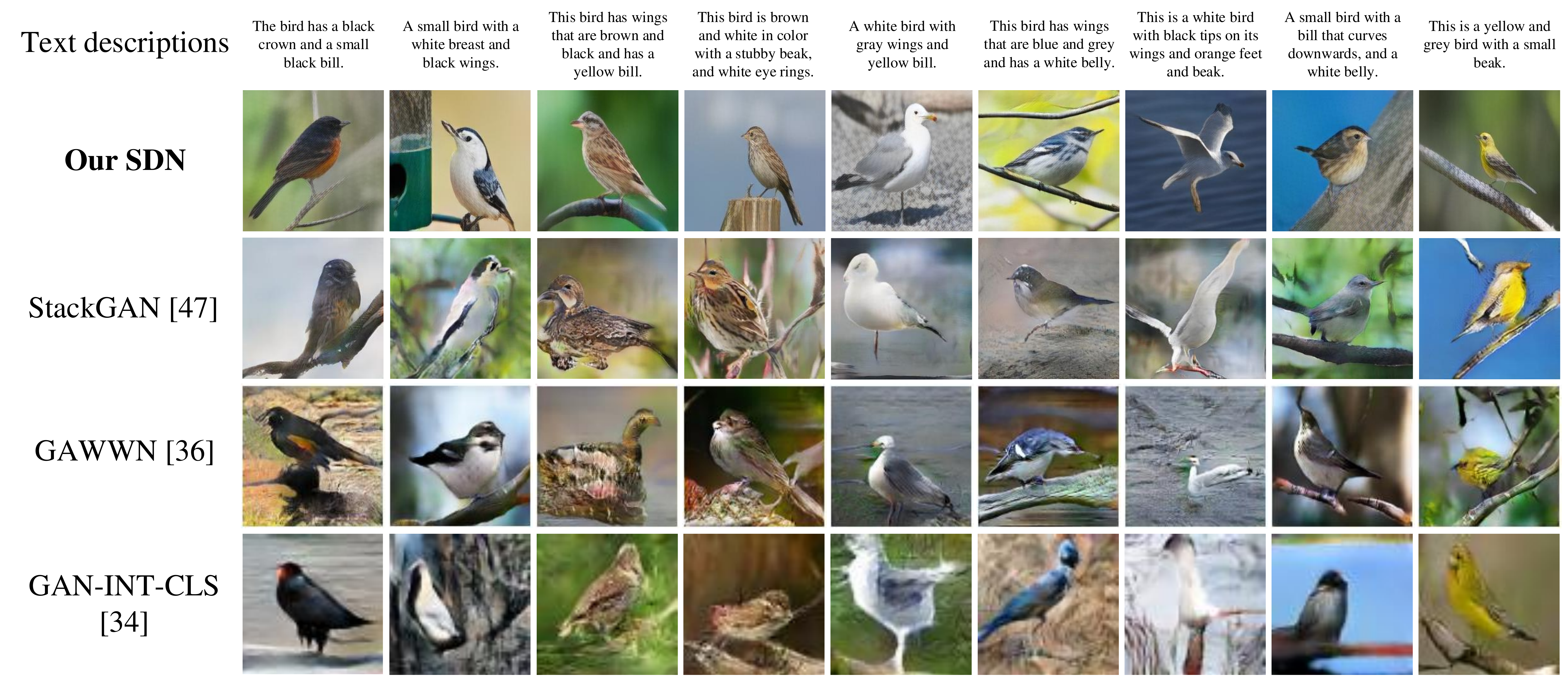}
	\end{center}
	\caption{Example results by our proposed SDN and compared methods on CUB-200-2011 testing set.}
	\label{fig:cub_example}
\end{figure*}

\begin{figure*}
	\begin{center}
		\includegraphics[width=0.85\linewidth]{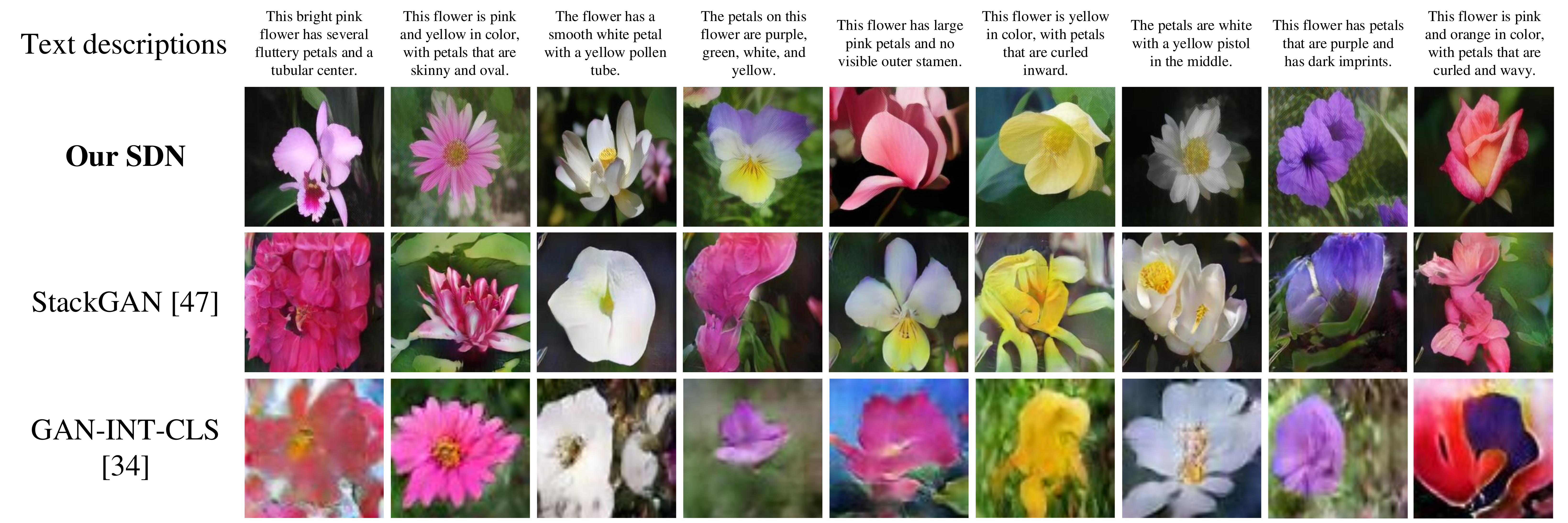}
	\end{center}
	\caption{Example results by our proposed SDN and compared methods on Oxford-Flower-102 testing set.}
	\label{fig:oxford_example}
\end{figure*}

\subsection{Qualitative Evaluation}

For qualitative evaluation, some text descriptions are randomly selected from the testing set. For each compared method, we generate 16 images conditioned on each text description and select the best one for comparison. Our SDN also synthesizes corresponding images conditioned on the same text descriptions.

Compared results on CUB-200-2011 dataset can be seen in Figure~\ref{fig:cub_example}. The images generated by GAN-INT-CLS have correct global structures conditioned on text descriptions, but fail to express the local details in most cases, such as vivid parts. Although GAN-INT-CLS uses deep symmetric structured joint embedding to relieve the heterogeneous gap, its synthetic images lose much necessary visual information in detail due to the simple heterogeneous correlation constraints. The GAWWN improves this situation by using object and part keypoint location constraints and additional conditioning variables. However, for fair comparison, we limit the GAWWN to generate images from text descriptions alone without additional location annotation, which is the same as the other methods. The images generated by GAWWN only conditioned on text descriptions still lack the authenticity in details. StackGAN can produce more realistic images with more visual information from text descriptions than the above approaches. Nevertheless, limited by first training phase of StackGAN, the structures of final synthetic images have some unrealistic characteristics. 	

Our SDN relieves these above problems and synthesizes images with correct global structures, local details and necessary visual information from text descriptions, which bridges the homogeneous and heterogeneous gaps. The same trends can be seen in the compared results on Oxford-Flower-102 dataset in Figure~\ref{fig:oxford_example}. 

\subsection{Diverse Synthesis}
\begin{figure}[!htb]
	\begin{center}
		\includegraphics[width=1\linewidth]{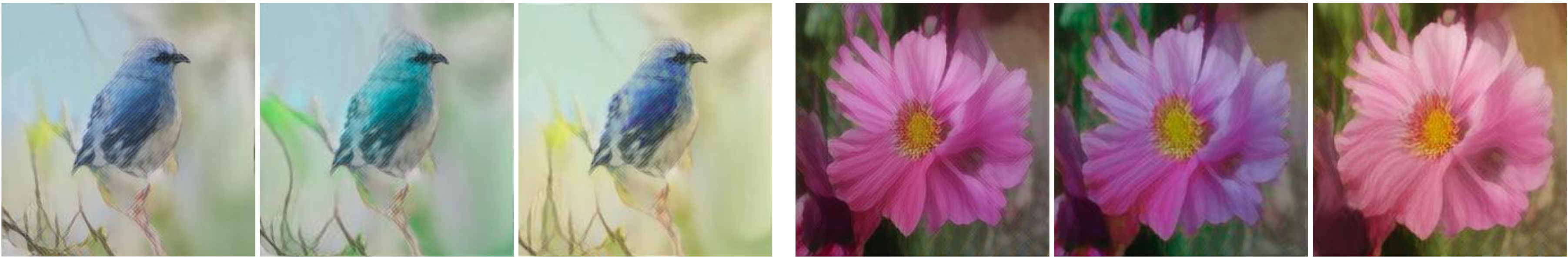}
	\end{center}
	\caption{Diverse synthesis examples of two input text descriptions.}
	\label{fig:diverse}
\end{figure}

Given a text description embedding as input, our SDN can generate diverse images as output. In our experiment, we fix the channels of $G$ final layer as $3$ to form an image for experimental convenience. If we change the output channels from $3$ to $3n$ in Stage-II distillation, SDN can generate $n$ diverse images for one text description. Some example results of diverse synthesis are shown in Figure~\ref{fig:diverse}. There some obvious difference in colors and subtle differences in shapes between these example results.

\subsection{Component Analysis}

\begin{figure*}
	\begin{center}
		\includegraphics[width=0.85\linewidth]{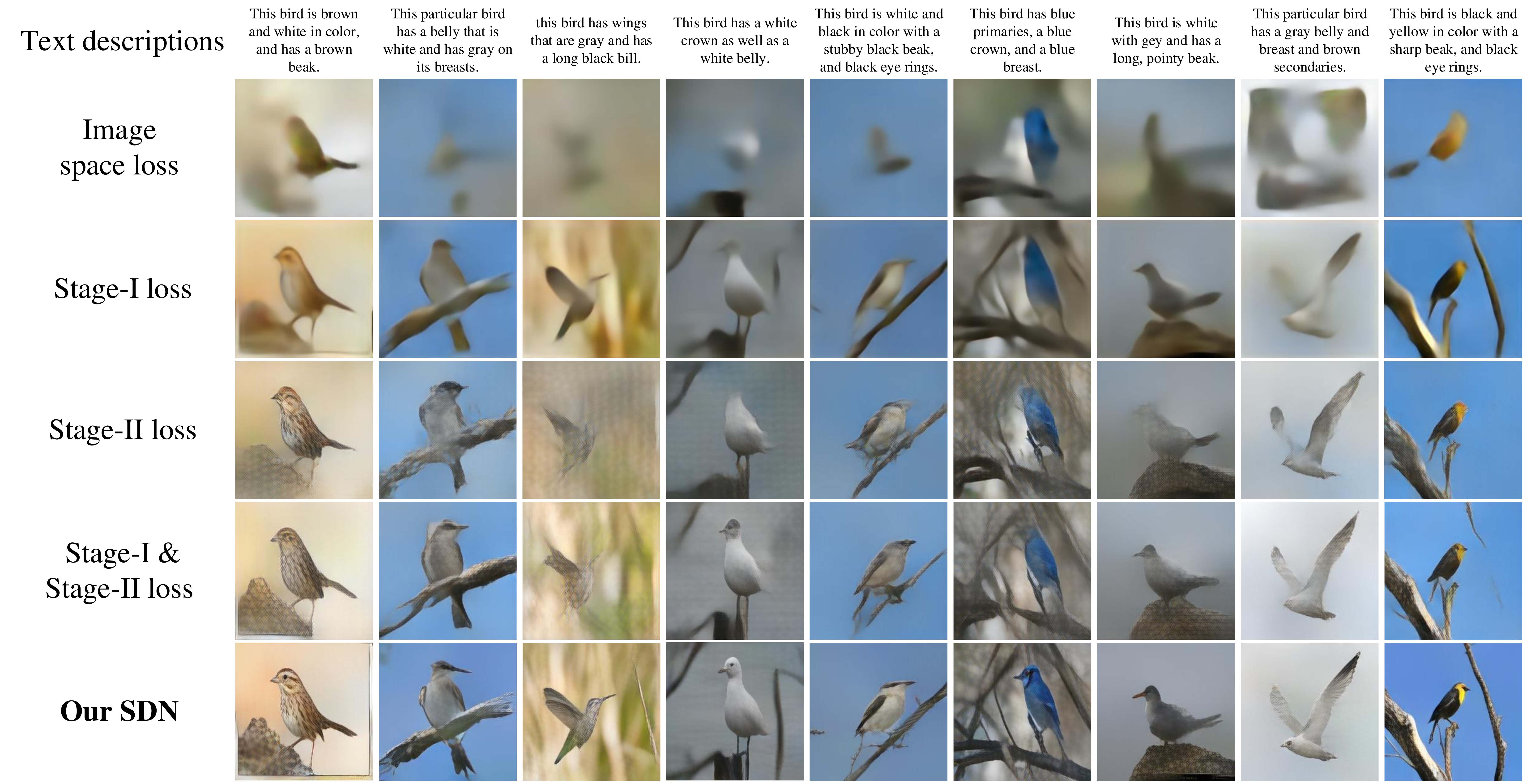}
	\end{center}
	\caption{Example results by our proposed SDN and the baselines on the CUB-200-2011 testing set.}
	\label{fig:baseline_example}
\end{figure*}

We conduct several baseline experiments on CUB-200-2011 dataset to analyze the effectiveness of each component in SDN. All the baseline schemes have the same network architecture as SDN but different loss function in training phase. The inception scores of these baseline schemes are shown in Table~\ref{table:Baseline}. Some representative examples of the experimental results are shown in Figure~\ref{fig:baseline_example}, which are introduced and analyzed as follows:

\begin{itemize}
	\item {\bf Image space loss}. 
	Image space loss is $Loss_{Image}$ formulated in Equation (2), which can guide the generative model to map the low-level feature space to image pixel space without distillation from source model. In Figure~\ref{fig:baseline_example}, we can see that the images generated by generative model only using the image space loss are excessively blurry. So only using image space loss cannot generate photo-realistic images conditioned on text descriptions due to the huge image space.
	
	\item {\bf Stage-I loss}. 
	This scheme only uses the Stage-I loss $Loss_I$ as Equation (1), which represents the primary guidance from source model to target model. Compared with image space loss in Table~\ref{table:Baseline} and Figure~\ref{fig:baseline_example}, we can see that Stage-I loss achieves better results obviously. However, there are still homogeneous and heterogeneous gaps in local details reflected by these images.
	
	\item {\bf Stage-II loss}. 
	``Stage-II loss'' corresponds to the $Loss_{II}$ as Equation (4), which is responsible for leading the generative model to learn more about the tiny discrepancy between synthetic and real images in multiple levels and capture integral text information. By comparing with Stage-I loss, we can see the Stage-II loss focuses on the local details and ignores the global structure of each image.
	
	\item {\bf Stage-I \& Stage-II loss}. 
	Considering the	shortcoming of the above schemes, it is reasonable to combine them in order to exploit their respective advantages. So the ``Stage-I \& Stage-II loss'' is an integration of these loss functions, simultaneously guiding the training process of target model as constraints. However, the Stage-I loss and Stage-II loss concentrate in the different aspects, which should have an order from global structure to the local details. 
	
	\item {\bf Diverse synthesis loss}. 
	We also evaluate the SDN with diverse synthesis loss in Equation (5) and (6) quantitatively and compare it with other schemes. 
	
\end{itemize}

\begin{table}[htb]
	\caption{Inception scores of the baselines and our SDN on the CUB-200-2011 testing set.}
	\begin{center}
		\begin{tabular}{|c|c|c|} 
			\hline
			Schemes & Inception scores\\
			\hline
			Image space loss & 2.11 $\pm$ 0.01 \\
			Stage-I loss & 3.90 $\pm$ 0.05 \\
			Stage-II loss & 4.77 $\pm$ 0.05 \\
			Stage-I \& Stage-II loss & 5.98 $\pm$ 0.07 \\
			Diverse synthesis loss & 6.84 $\pm$ 0.05 \\
			\textbf{our SDN} & \textbf{6.89 $\pm$ 0.06} \\
			\hline
		\end{tabular} 
	\end{center}
	\label{table:Baseline}
\end{table}

By numerically analyzing the inception scores in Table~\ref{table:Baseline} and visually inspecting the generated samples in Figure~\ref{fig:baseline_example}, we can conclude that our SDN adopts a very effective scheme and finally produces images with admirable quality conditioned on text descriptions. Moreover, the SDN with diverse synthesis loss can also achieve an effective performance.

\subsection{Approximate Sample Retrieval}

\begin{figure}[t]
	\begin{center}
		\includegraphics[width=1\linewidth]{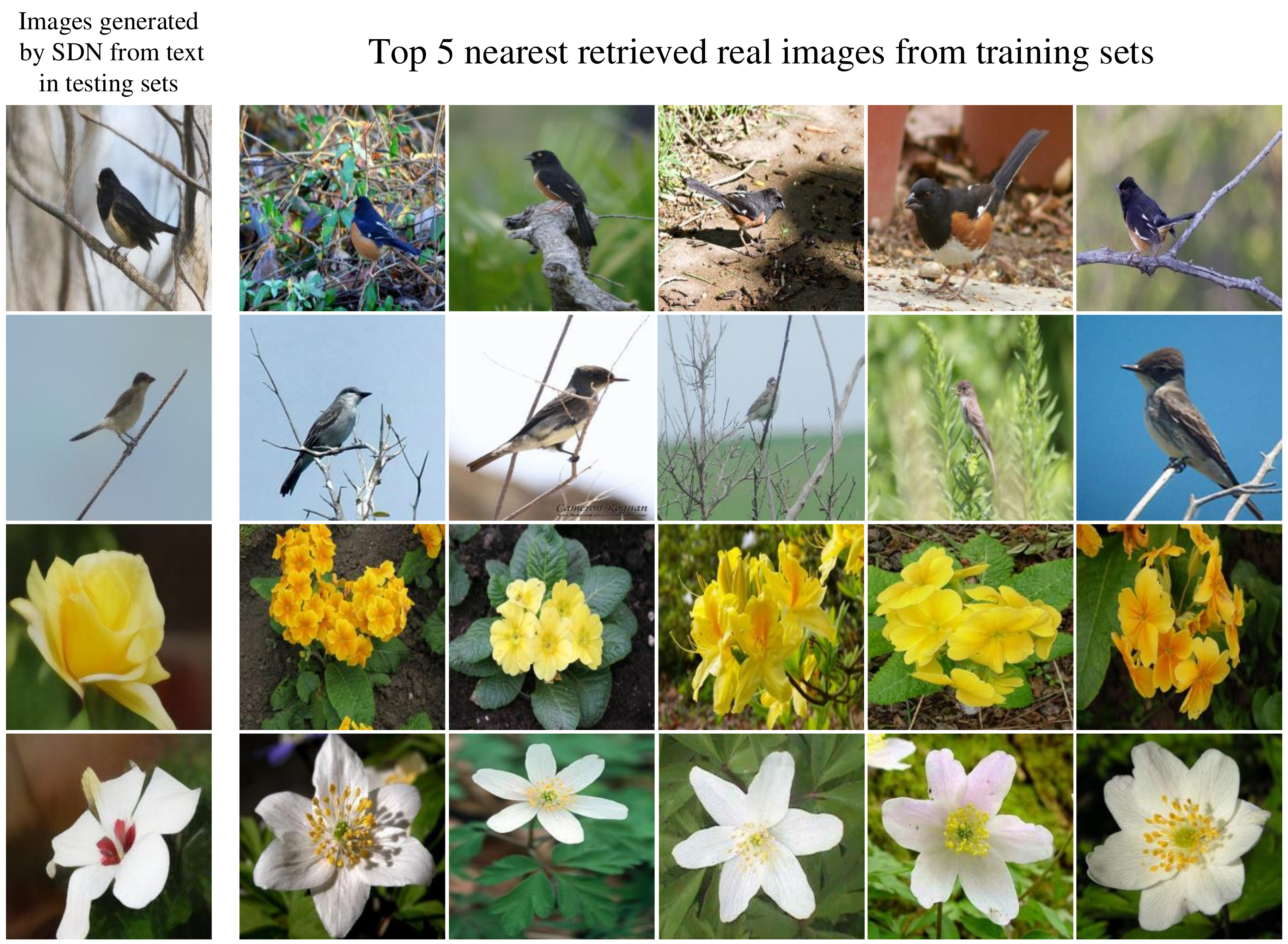}
	\end{center}
	\caption{For images generated by SDN (column 1), we retrieve their top-5 training images (columns 2-6).}
	\label{fig:retrieval}
\end{figure}

To verify that our SDN has learned the distribution of real images and the heterogeneous correlation instead of imitating the appearance of training samples, we conduct this approximate sample retrieval. We fine-tune the VGG19 model on CUB-200-2011 or Oxford-Flower-102 training set and use it to extract fc7 layer visual features for images generated by SDN and all the images in training set. Then we retrieve all the images in the training set with cosine similarity for the synthetic images and select the top-5 samples as the nearest neighbors to present. There are four top-5 retrieved results of synthetic images on CUB-200-2011 and Oxford-Flower-102 training sets demonstrated in Figure~\ref{fig:retrieval}. It is obvious that there is some definite difference on key characteristics between the synthetic images and retrieved training samples, although they are similar in colors or shapes. For example, on the third and fourth row in Figure~\ref{fig:retrieval}, even the retrieved flower images have similar color with a synthetic image query, their shapes are obviously different. In other word, SDN creates its unique flowers conditioned on the text descriptions, which are unseen in the training set.

\subsection{Sentence Interpolation}

\begin{figure}[t]
	\begin{center}
		\includegraphics[width=1\linewidth]{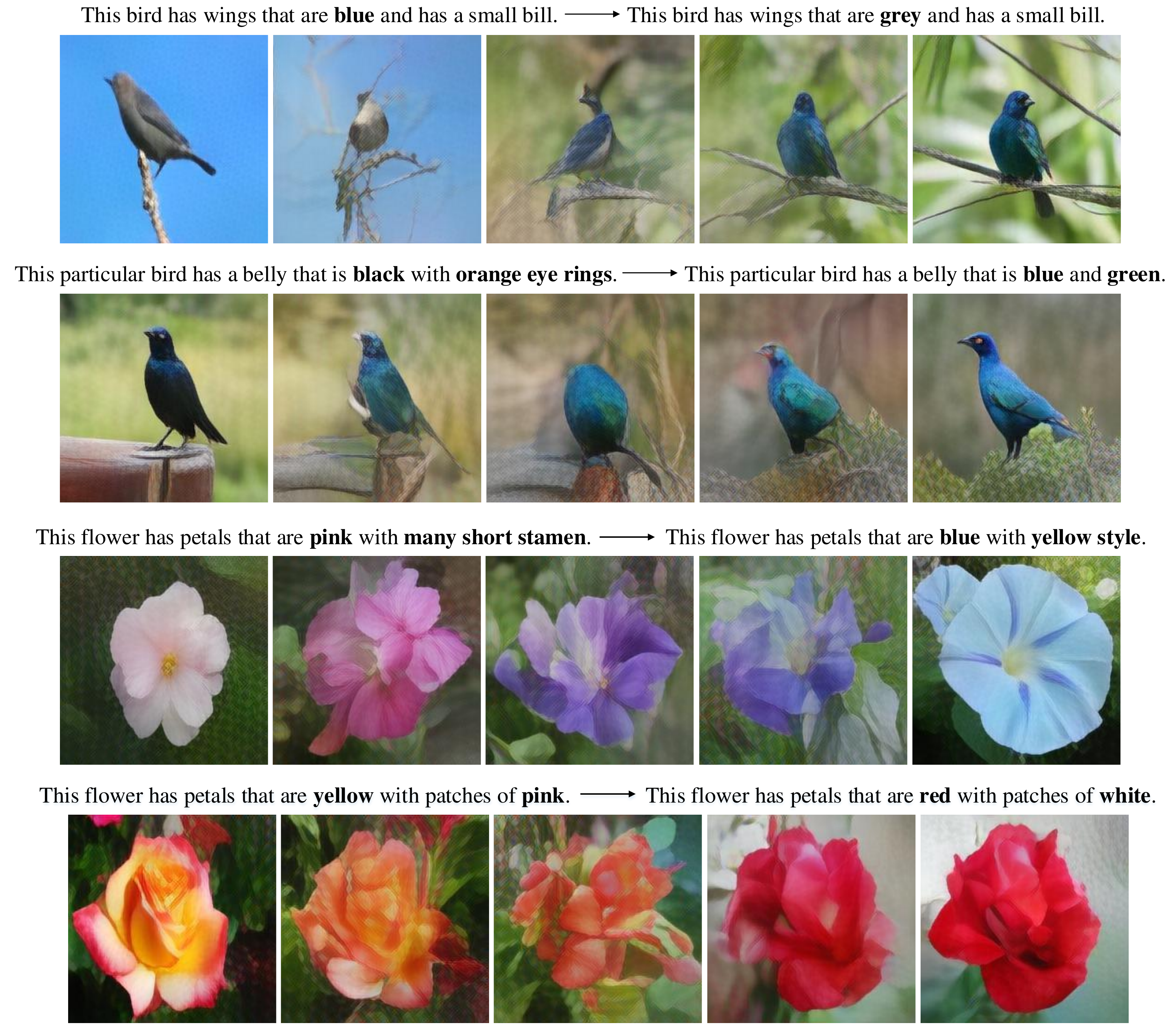}
	\end{center}
	\caption{Generated bird and flower images by interpolating two sentences are shown here from left to right.}
	\label{fig:change}
\end{figure}

We present the learned latent manifold by a sentence interpolation experiment. We use linearly interpolated sentence embeddings to generate corresponding images and show four examples in Figure~\ref{fig:change}. Because the intervening points do not have correct ground-truth text descriptions, the interpolated images generated by SDN lack authenticity in details. But some visual characteristics of images generated by SDN also can be distinguished such as the gradient color in all the four rows of Figure~\ref{fig:change}. And this is a verification that the synthetic images reflect the meaning changes of sentences. For example, the basic color of flower in fourth row is changed from yellow to red. Moreover, the same trend can be found in local details. In the third row, the petal shape of first left flower is different with the first right flower. So as we can see, the petal shape on second right images is unique with these two kinds of flowers, which is compromised as an intervening point. We can conclude that our SDN learns a smooth heterogeneous manifold not only in global structure but also in local details due to the hierarchical knowledge distillation from generic discriminative model.

\subsection{Failure Cases}

\begin{figure}[t]
	\begin{center}
		\includegraphics[width=1\linewidth]{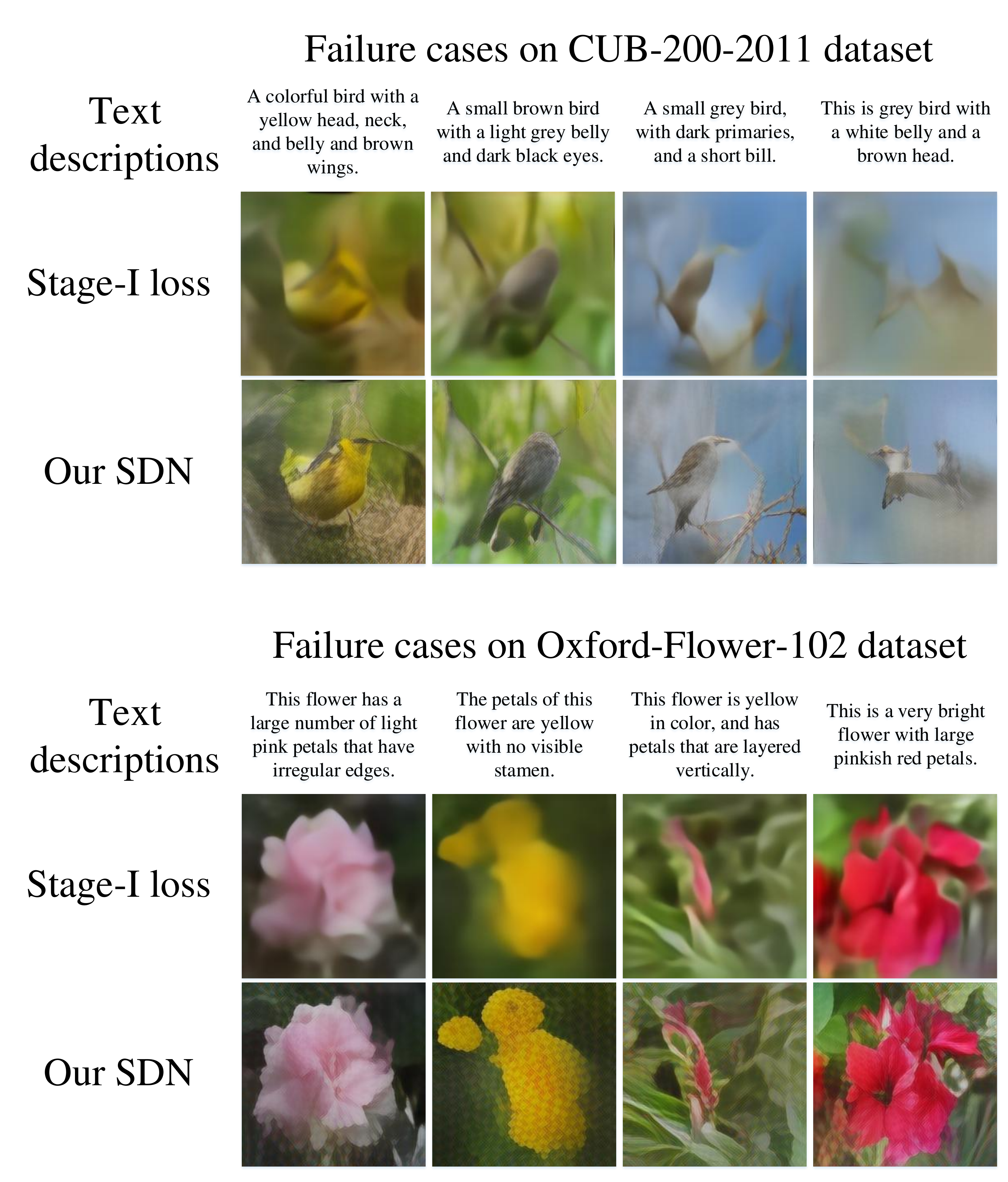}
	\end{center}
	\caption{Failure cases on CUB-200-2011 and Oxford-Flower-102 datasets.}
	\label{fig:failure}
\end{figure}

We present some failure cases on the two datasets in Figure~\ref{fig:failure}. As them shown, these failure cases are mainly caused by the uncertain object shapes produced by the Stage-I loss scheme. Although the main colors are basically correct of these failure cases, their object shapes are wrong due to the limitation of Stage-I distillation. The knowledge of source discriminative model is so generic that the target generative model cannot learn specific representations well for some text descriptions. We will consider and deal with this problem in future work.

\section{Conclusion}
In this paper, we have proposed Symmetrical Distillation Networks (SDN) for addressing the problem of heterogeneous and homogeneous gaps in the text-to-image synthesis task. Our SDN consists of a source discriminative model and a target generative model which has a symmetrical structure with the source model and is trained with a two-stage distillation paradigm. By using this symmetrical structure and training paradigm, the target model can be trained finally and synthesize images with multi-level characteristics.

In the future, we plan to extend our work in two aspects. On one hand, we intend to apply SDN to other visual information generation applications such as human pose image or video synthesis to verify its generality. On the other hand, we will focus on increasing the resolution of synthetic images by more powerful discriminative models.

\section*{Acknowledgments}
This work was supported by the National Natural Science Foundation of China under Grant 61771025 and Grant 61532005.

\clearpage
\bibliographystyle{ACM-Reference-Format}
\balance
\bibliography{sample-bibliography}

\end{document}